\pdfoutput=1

\documentclass[11pt]{article}

\usepackage{acl}

\usepackage{tipa}

\usepackage{epsfig}
\usepackage{epstopdf}

\usepackage{caption}
\usepackage{subcaption}

\usepackage{hyperref}

\usepackage{amsmath}

\usepackage{cleveref}
\crefname{section}{\S}{\S\S}
\Crefname{section}{\S}{\S\S}
\crefname{table}{Table}{}
\crefname{figure}{Fig.}{Figs.}
\crefname{algorithm}{Algorithm}{}
\crefname{algorithm}{Algorithm}{}
\crefname{line}{Line}{}
\crefname{appendix}{App.}{}
\crefformat{section}{\S#2#1#3}
\crefname{thm}{Theorem}{}
\crefname{cor}{Corollary}{}
\crefname{prop}{Proposition}{}
\crefname{def}{Definition}{}

\usepackage[disable]{todonotes}
\makeatletter
\newcommand*\iftodonotes{\if@todonotes@disabled\expandafter\@secondoftwo\else\expandafter\@firstoftwo\fi}  
\makeatother

\usepackage{times}
\usepackage{latexsym}

\usepackage[T1]{fontenc}

\usepackage[utf8]{inputenc}

\usepackage{microtype}

\usepackage{multirow}

\title{Word Order Matters when you Increase Masking}

\newcommand{\lattice}{\normalfont \text{\textipa{@}}}
\newcommand{\unipi}{\normalfont \text{\textipa{B}}}

\author{
Karim Lasri$^{\lattice,\unipi}$ Alessandro Lenci$^{\unipi}$ Thierry Poibeau$^{\lattice}$ \\
$^{\lattice}$Lattice (\'Ecole Normale Supérieure-PSL, CNRS, U. Sorbonne Nouvelle)~\; \\ 
~$^{\unipi}$University of Pisa~\;  \\
  \texttt{\href{mailto:karim.lasri@ens.psl.eu}{karim.lasri@ens.psl.eu}}~\;~\\
  \texttt{\href{mailto:alessandro.lenci@unipi.it}{alessandro.lenci@unipi.it}}~\;~ \texttt{\href{mailto:thierry.poibeau@ens.psl.eu}{thierry.poibeau@ens.psl.eu}}
}

\begin{document}
\maketitle
\begin{abstract}
Word order, an essential property of natural languages, is injected in Transformer-based neural language models using position encoding. However, recent experiments have shown that explicit position encoding is not always useful, since some models without such feature managed to achieve state-of-the art performance on some tasks. To understand better this phenomenon, we examine the effect of removing position encodings on the pre-training objective itself (i.e., masked language modelling), to test whether models can reconstruct position information from co-occurrences alone. 
We do so by controlling the amount of masked tokens in the input sentence, as a proxy to affect the importance of position information for the task. We find that the necessity of position information increases with the amount of masking, and that masked language models without position encodings are not able to reconstruct this information on the task. 
These findings 
point towards a direct relationship between the amount of masking and the ability of Transformers to capture order-sensitive aspects of language using position encoding.


\end{abstract}

\maketitle

\section{Introduction}
Transformer-based language models have become ubiquitous since they demonstrated improvements in most NLP downstream tasks \citep{devlin-etal-2019-bert, liu-roberta-2019}. A lot of ink has been spilled regarding the amount of linguistic structure that such models captured \citep{jawahar-etal-2019-bert}, pointing towards the acquisition of diverse linguistic abilities. 
As neural language models need to process information about the position of their input tokens to capture structural generalizations, a plethora of proposals to encode such information have been made \citep{press, he-liu, su-lu, chang-etal-2021-convolutions, chen-etal-2021-simple, chen-2021-permuteformer}. Recent work, however, questioned whether word order information is really useful for pre-trained models to solve downstream tasks \citep{sinha-etal-2021-masked}, showing that models could perform well when using only higher-order co-occurrence statistics. 
Other authors \citep{haviv} have shown that some transformers could reconstruct partly position information without it being explicitly injected. Examining performance on downstream tasks can show that the task simply does not require order information
, or that the dataset used to test the model is too easy \citep{abdou-etal-2022-word}, leading to indirect observations regarding a model's ability to reconstruct position information. 

In turn, we choose to test the importance of position encodings for the pre-training task itself, masked language modeling, to get more direct evidence about whether and when position matters to language models. 
We do so under different amounts of masking, as intuitively, position information should be increasingly important when more tokens are missing from the context. 

Our experiments show that when masking only one token, the absence of position encoding has little effect on the model's performance. However, its importance increases with the number of masked tokens, forcing the model to leverage position information to perform better on its training objective. This finding should draw our attention towards choosing more carefully the amount of masking to train masked language models -- a choice as important as the position encoding scheme itself. 


\section{Related work}
A recent line of research investigated the importance of word order information during pre-training for models to solve downstream tasks, showing little variations when their input sentences are shuffled \citep{pham-etal-2021-order, sinha-etal-2021-masked, hessel-schofield-2021-effective}. In a similar line of research, \citep{haviv} found that even in absence of position encoding, models were still able to reconstruct the latter when probed for tokens' absolute position information in their intermediate layers. This finding in turn questioned the need for injecting explicitly position information in language models. \citep{abdou-etal-2022-word} also showed that shuffled models were still able to capture position information even when information about word order was removed after subword segmentation, likely because of the dependency between unigram occurrence probability and sentence length.
Given all this work, it is surprising that the importance of explicit word order information in a neural language model still eludes us. 
In this study, we choose to investigate more carefully this phenomenon, and propose a methodology carefully designed to evaluate the importance of position encoding for the pre-training objective. 

\section{Experimental Setup}

\subsection{Methodology}

In our experiments, the goal is to investigate the extent to which a transformer neural model requires explicit position encoding to perform well on the masked language modeling objective. We do so under different amounts of masking to examine how this parameter affects the need for explicit position encoding. We make use of two variants for each trained model, one in which we inject position information, and one deprived from explicit access to that information. To evaluate whether the trained model reconstructs its input sentences using position information, we compare its probability estimates $q$ to two versions of the language's true probabilities $p$ on the validation set. The first version, $p_{o}$, represents the probability of completions given the original, ordered input context. The second version, $p_{u}$, is the probability given unordered contexts. In the following sections, we explain how we perform this comparison to evaluate the extent to which explicit position encoding is required for the masked language modeling task. 


\subsection{Data}
\label{sec:data}
\begin{figure}
     \begin{subfigure}[b]{0.48\textwidth}
         \centering
         \includegraphics[width=\linewidth]{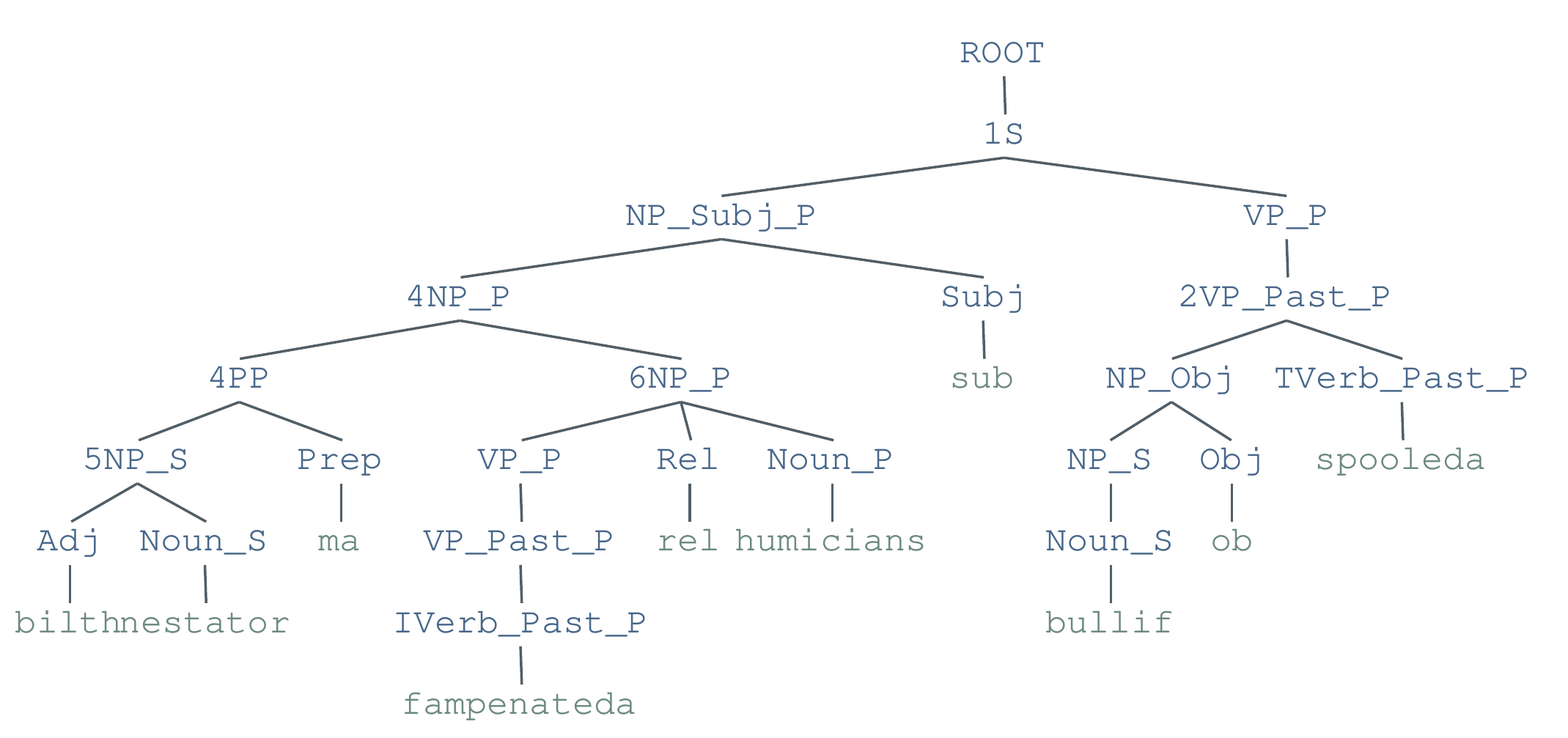}
     \end{subfigure}%
     \centering
     \hspace{5pt}
     \begin{subfigure}[b]{0.48\textwidth}
         \centering
         \includegraphics[width=\linewidth]{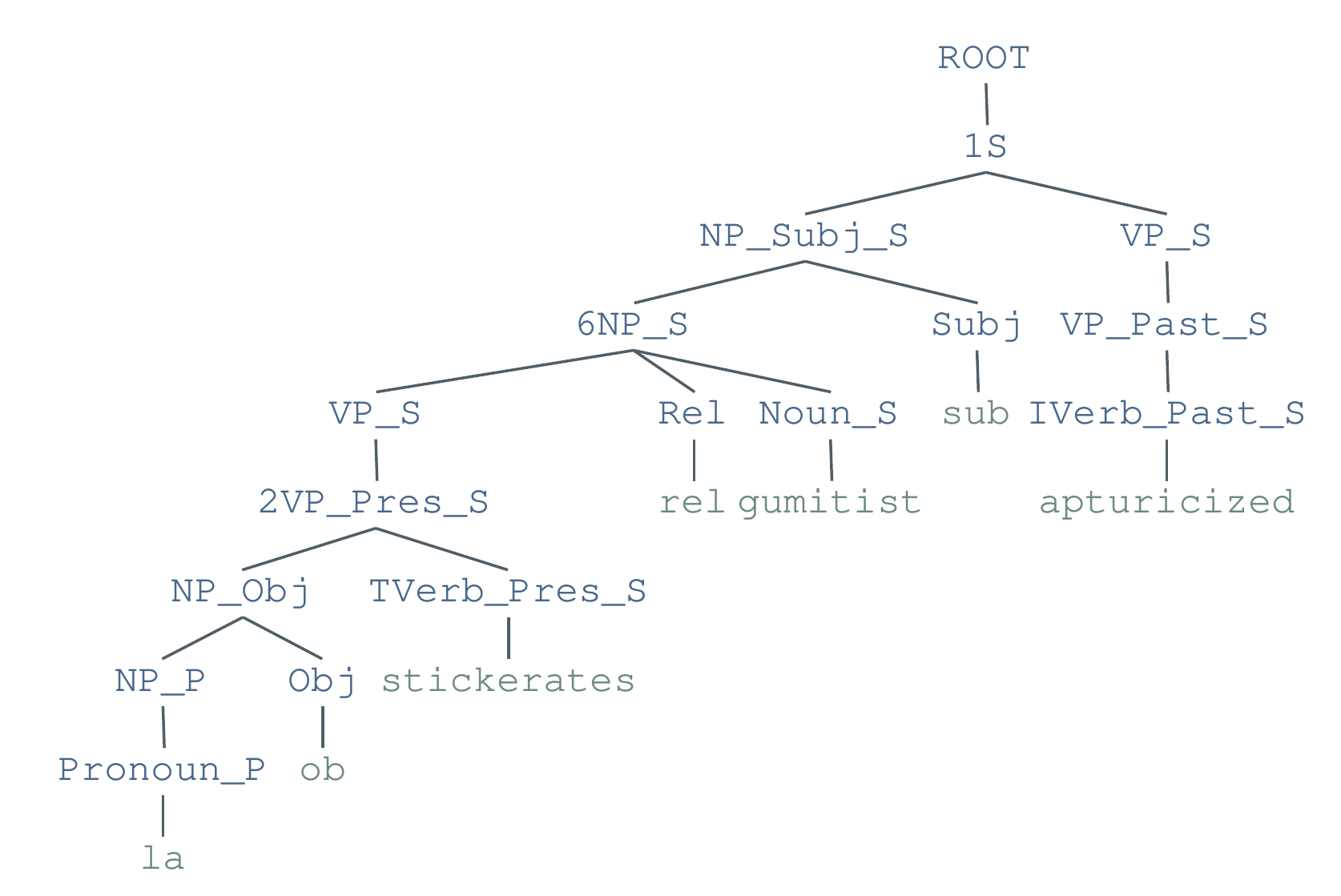}
     \end{subfigure}
    \caption{Examples of sentences found in the artificial language used for our analysis.}
    \label{fig:trees}
    \vspace{-5pt}
\end{figure}

When using natural languages, it is hard to assess whether the model indeed relies on order information because it is not easy to design a dataset controlled to target specifically the usage of position information. In particular, as one does not have access to the true probability distribution of natural languages, it is hard to make clear predictions regarding how a model not using position information should behave. On the other hand, artificial languages obtained from a generative procedure that is known \textit{a priori} make it possible to get tight estimates of their true probability distribution, both with, and without access to position information. The use of artificial languages has sparked interest over the past years, as a proxy to test targeted properties of neural models in controlled settings \citep{white-cotterell-2021-examining, wang-eisner-2016-galactic}.
In our experiments, we make use of data released by \citet{white-cotterell-2021-examining}.
The dataset consists of sentences generated from an artificial grammar, using a CFG such that all production rules have fixed probabilities.\footnote{The artificial language features certain constraints present in natural languages such as morphological agreement relations.} 
This design makes it possible to evaluate the true probability of completions given masked input sentences, as a comparison point to the model's observed behavior.
We display examples of generated trees in \cref{fig:trees}.\footnote{In our experiments, we used the unaffected artificial language (Grammar 000000) released by \citet{white-cotterell-2021-examining}. The original code can be found at \url{https://github.com/rycolab/artificial-languages}.}

\subsection{Estimating the true probability distribution of the task}
We exploit our direct access to the generative procedure which produces our input sentences to estimate the true probability distribution of the masked language modeling task. 
We do so by assuming that the context is either ordered or not.
Specifically, we generate sequences recursively using the artificial language's production rules, until the probability sum of fully expanded sentences\footnote{i.e. sequences that have no non-terminal label.} reaches a certain coverage.\footnote{We generate sentences along with their true sentence probability in our artificial language until we reach a probability sum superior to 0.75} We then iterate over these sentences to mask words at each position and aggregate completions for sequences that share the same unmasked context in the Masked Language Modelling setting. We thus obtain a probability distribution of completions $Y$ given (ordered) masked contexts $X_{o}$, which we write $\{p_{o}(y|x) \mid x,y \in X_{o}\times Y_{o} \}$.\footnote{Note that when using natural languages, automatic collection of sentences in real corpora does not allow access to all possible completions in context, in addition to only providing sparse, and often biased, samples of sentences. Thus the true probability remains unknown, as noted in \cref{sec:data}.}

We also compute a second version of the probability distribution that assumes no ordering of the context, aggregating completions for unmasked sequences whose unordered masked context is the same in $X_{u}$, obtaining $\{p_{u}(y|x) \mid x,y \in X_{u}\times Y_{u} \}$. To get the probability for unordered contexts, we simply group input sequences by sorting their elements alphabetically to remove order information and sum their probabilities for each unordered context. As we only use this procedure to remove information when estimating the task's true probability, the inputs which are seen by our models remain unchanged. As this removes all word order information when estimating the MLM task's probability distribution, our estimate is only dependent on information about each token's number of appearances in each input.

\subsection{Is position information necessary for the task ?}
\label{sec:necessity}
Given the true probabilities $p_{o}$ and $p_{u}$ for our task, we want to measure how different these are. We compute the KL-divergence :
\begin{equation}
    \label{eq:information-task-divergence}
    D_{KL}(p_{o}, p_{u}) = \sum_{x,y\in X_{o}\times Y_{o}}p_{o}(y|x) \log \frac{p_{o}(y|x)}{p_{u}(y|x)}
\end{equation}

This statistical distance allows us to estimate how different are the two distributions. We predict that by masking more tokens, the task would increasingly require position information and the divergence would also increase.\footnote{Note that while the KL-divergence is asymmetric, in this order the quantity represents the information gain achieved by having access to position information.} 

\subsection{Is position encoding useful to the model?}
\label{sec:models}
We test two variants of the BERT architecture \citep{devlin-etal-2019-bert}, using Huggingface's Transformer library \citep{wolf-etal-2020-transformers}. In the first model, position information is encoded using learned absolute position embeddings,\footnote{This encoding scheme is widespread in transformer-based models, see \citet{dufter-position-overview-2021} for an overview} while such explicit encoding is removed from the second. We call such models \textbf{BERT} and \textbf{NP}.
Their hyperparameters are described in \cref{app:model-hyperparameters}.
For each model, we compare its probability estimates $q$ in context to the task's true distribution assuming both that position information is present in contexts $p_{o}$, and absent $p_{u}$. We do so by computing the KL-divergence between $q$ and $p \in (p_{o}, p_{u})$ as follows:
\begin{equation*}
D_{KL}(p,q) = H(p,q)-H(p)
\end{equation*}
We estimate the true entropy $H(p)$ for the masked language modeling (MLM) task using either $p_{o}$ or $p_{u}$ on our set of generated sentences:

\begin{equation}
\begin{split}
H(Y|X) & = - \sum_{x,y \in X\times Y} p(x,y) \log\frac{p(x,y)}{p(x)} \\
 & = - \sum_{x,y \in X \times Y} p(y|x) p(x) \log p(y|x)
\end{split}
\end{equation}
For each context, we compute the true entropy of its completions :
$$\forall x \in X \text{, }h_{Y}(x)= - \sum_{y \in Y}p(y|x)\log p(y|x)$$
And we finally compute the task entropy by averaging these context entropies over our kept masked contexts $X_{o}$ or $X_{u}$ :
$$H(Y|X) = \sum_{x \in X} p(x) h_{Y}(x)$$

\noindent{}We obtain two true task entropy estimates, $H(p_{o})$ for ordered contexts, and $H(p_{u})$ for unordered ones. For each model, we then estimate the cross entropy to each true distribution. Denoting the model's output probability $q$, the cross-entropy writes as follows\ :

$$H(p,q)=-\sum_{x,y \in X \times Y} p(y|x)\log q(y|x)$$

\noindent{}We then use the task's true entropy and the model's cross-entropy to compute the KL-divergence. For each model, by comparing $D_{KL}(p_{o}, q)$ to $D_{KL}(p_{u}, q)$, we can assess whether the model's estimates fit better the task's probability for ordered contexts, or unordered contexts.
If explicit position encoding is necessary, we predict that  $D_{KL}(p_{u}, q)$ should be greater than $D_{KL}(p_{o}, q)$ for \textbf{BERT}, and lower for \textbf{NP}. Otherwise, both models should have similar behavior.

\subsection{Testing the effect of masking}
In this study, we compare \textbf{BERT} and \textbf{NP} under different amounts of masking. We surmise that increasing that parameter should increase the necessity of using position information, as measured by \cref{eq:information-task-divergence}. If this is the case, varying this parameter will allow us to investigate whether position encoding is necessary as the task increasingly requires using that information. 


\begin{figure}[h]
     \centering
     \includegraphics[width=\linewidth]{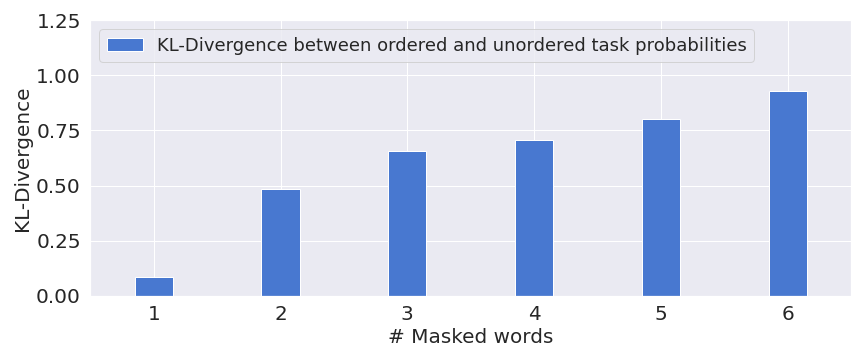}

    \caption{KL-Divergence between the true task probabilities assuming ordered and unordered inputs.}
    \label{fig:kldiv-data}
    \vspace{-5pt}
\end{figure}

\section{Results}
We first display the KL-divergence between true probability distributions assuming ordered and unordered contexts in \cref{fig:kldiv-data}. In accordance with our expectations,\footnote{see \cref{sec:necessity}} when increasing the amount of masking, the true distribution of completions given ordered contexts diverges from that of unordered contexts. Interestingly though, when only one token is masked, the divergence is low. This suggests that in this setting, models should have little difference regardless of whether they have access to explicit position information. By increasing the amount of masked tokens, we can further observe that the two considered true probabilities $p_{o}$ and $p_{u}$ diverge. We thus expect that models should increasingly rely on position information to approximate the true ordered distribution.

\begin{figure}
     \begin{subfigure}[b]{0.48\textwidth}
         \centering
         \includegraphics[width=\linewidth]{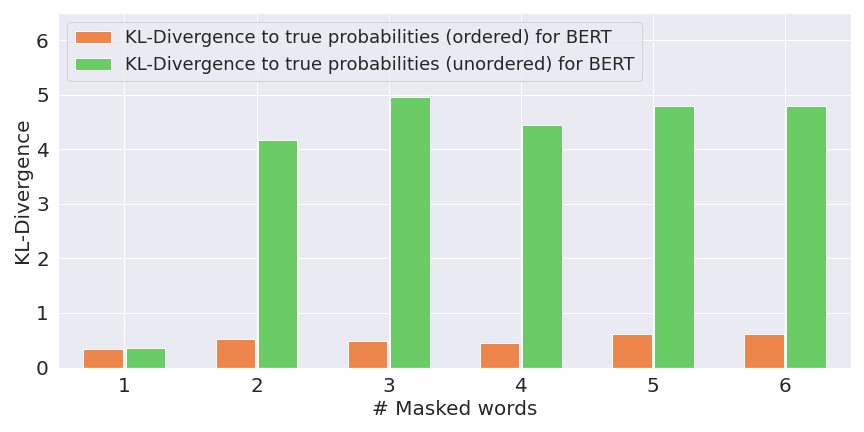}
     \end{subfigure}%
     \centering
     \hspace{5pt}
     \begin{subfigure}[b]{0.48\textwidth}
         \centering
         \includegraphics[width=\linewidth]{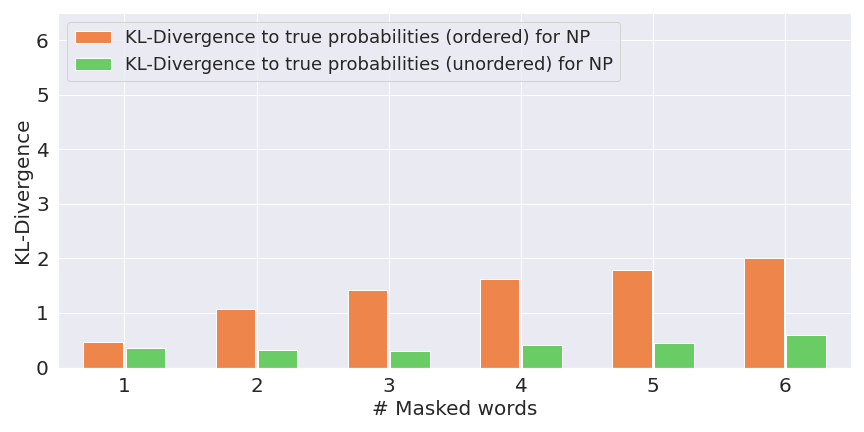}
     \end{subfigure}
    \caption{KL-Divergence between the true task probabilities and our models' probability estimates (\textbf{BERT}-top and \textbf{NP}-bottom), assuming contexts are ordered (orange bars) and unordered (green bars).}
    \label{fig:kldiv-model-data-results}
    \vspace{-5pt}
\end{figure}

We further display how well each model approximates each probability estimate in \cref{fig:kldiv-model-data-results} to verify whether the presence of position encoding is useful to the masked language modeling task under different amounts of masking.

Expectedly, the model with no position encoding scheme performs similarly to the BERT model when only one token is masked. In this setting, the context contains enough information for the model regardless of whether it sees its input tokens as ordered or as a bag of words. When masking more tokens however, this difference becomes increasingly marked.\footnote{See \cref{app:model-perplexities} for our models' perplexities.}

Further, we observe that the \textbf{BERT} model has a low divergence to the true probability assuming ordered contexts regardless of the amount of masking, while it diverges increasingly from the distribution that assumes no ordering of the context. The opposite pattern holds for the \textbf{NP} model. Taken together, these results show that position encoding is necessary to approximate the true distribution of the task when it requires position information, that is when the number of masked tokens is increased.

\begin{figure}[h]
     \centering
     \includegraphics[width=0.5\textwidth]{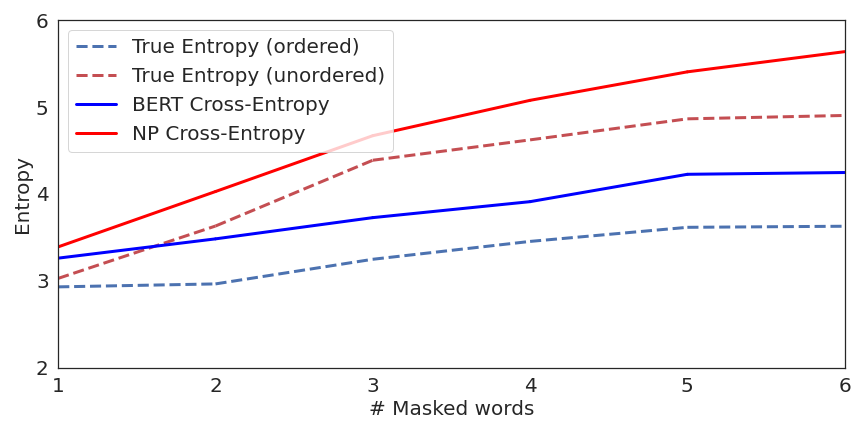}
     \caption{A comparison between entropies of true probabilities for the MLM task (assuming ordered and unordered contexts), and our models' cross-entropies}
     \label{fig:model-vs-data-entropies}
 \end{figure}

In \cref{fig:model-vs-data-entropies}, we compare our models' cross-entropies to the task's true entropies.
The figure aggregates the two main observations made in this article, that when the number of masked tokens increases : (i) the true entropy of the data with and without position diverge from each other, and (ii) that position encoding is required to approximate the task's true probability distribution assuming ordered contexts. 
Accordingly to our previous observations, the \textbf{NP} model, which does not have access to the ordering of tokens, has a cross-entropy that fits the true probability distribution's entropy assuming no ordering of the context (red lines).
Looking at \textbf{BERT}'s cross-entropy, we see that this model, which has access to position information, rather fits the true probability distribution assuming the context is ordered.

\section{Discussion}
\subsection{Position encoding and language modeling}
Previous work claimed that transformer autoregressive language models without position encodings could reconstruct position information by inferring the number of preceding tokens, but not bidirectional transformer models \citep{haviv}. Testing a RoBERTa model \cite{liu-roberta-2019} led to great difference in perplexity when removing position information at the input level. However, we show this difference to strongly depend on the amount of masking : as autoregressive language models predict only one token at a time, the task could be equally easy for models deprived from position information. Our results call for increased scrutiny when comparing autoregressive and masked language models, making sure that they are asked to predict comparable numbers of tokens.


\subsection{Mask more !}
In our study, we have shown that the utility of explicit position encoding increases with the number of masked tokens. This finding echoes 
\citet{wettig}'s study, showing that masking 40\% of tokens rather than 15\% during pre-training leads to better performance on downstream tasks. This evidence could draw more attention towards understanding how different amounts of masking can lead models to abstract away from position information, and capture more structural knowledge about the languages they model.

\section{Conclusion}
In this work, we evaluated the importance of position encoding for a masked language model. We showed that without explicit access to position information, a model can obtain performance similar to a model that learns position embeddings, when only one token is masked. We find that when increasing the number of masked tokens, the output probability distribution assuming unordered inputs diverges from that which assumes ordered sentences, reflecting that the task increasingly requires making use of position information. We further show that under this condition, models with position encoding outperform their counterpart deprived from position information. This in turn could draw more attention to the amount of masked tokens, which might be a crucial parameter for models to abstract away from their input sentences' position information, in addition to the chosen position encoding scheme.

\section{Limitations}

The results we have presented in this paper were obtained over artificial languages. Adapting the method to natural languages may be difficult. 
\paragraph{The true probability distribution is not accessible for natural languages.}
In this study, we investigate how the amount of masking impacts the usage of position encoding by a neural language model. We chose to carry out this experiment on an artificial language, because of the ease to access the true probability distributions in each setting. While this result informs us that the amount of masking could be key for masked language models to use and abstract away from position information extracted from their input, this methodology is not easy to adapt to natural languages, because the true probability distribution is not accessible for natural languages. In future work, one could try to find proxies to estimate reference points for natural languages, with potentially looser estimates than the one used in this study. 

\paragraph{Training several masked language models on natural languages is computationally expensive.}
In order to investigate how the amount of masking impacts the degree to which a NLM makes use of its position encodings, or higher-order structural properties of natural languages, one would need to train a large neural model for each condition under investigation, and for each retained amount of masking. This, added to the potential hyperparameter space search would require substantial computing resources as training a model on natural languages requires large amounts of data during training. 

\paragraph{Natural languages are usually more flexible regarding word order.}
In our experiments, we investigate the impact of masking on using position information using artificial languages where word order is fixed. We conclude that neural language models make use of position information on the masked language modeling objective when the number of masked tokens increases. However, while this should hold true for data similar to ours, where the word order is fixed and hence position information greatly affects which token needs to be predicted at a certain position, we cannot make claims regarding the impact of masking on languages where word order is more variable, which is the case of any natural language. Further analyses are needed to evaluate whether position encoding impacts language modeling in different ways when word order is rather fixed (like English), compared to when it is more variable (like in Latin or Finnish). 

\section*{Ethics Statement}
The authors foresee no ethical concerns with the work presented in this paper.

\section*{Acknowledgements}
This work was funded in part by the French government under management of Agence Nationale de la
Recherche as part of the ``Investissements d’avenir''
program, reference ANR-19-P3IA-0001 (PRAIRIE
3IA Institute).

\bibliography{anthology,custom}

\clearpage

\appendix

\section{Data Statistics}
\label{app:data}


We describe in \cref{tab:dataset-stats} some statistics for the dataset used to train our models. 

\begin{table}[h]
 \begin{small}
 \begin{center}
 \begin{tabular}{ |l|c| } 
 \hline
  \textbf{Train size} & 100000 \\
  \textbf{Test \& Validation Size} & 10000 \\
  \textbf{Vocabulary Size} & 1261 \\
  \textbf{Mean Sentence Length} & 12.51 \\
 \hline
 \end{tabular}
 \caption{\label{tab:dataset-stats} Statistics of the dataset used to train our models.}
 \end{center}
 \end{small}
 \end{table}

\section{Model Hyperparameters}
\label{app:model-hyperparameters}
The architectures' hyperparameters are common to both our \textbf{BERT} and \textbf{NP} models.
The learned tokenizer has been trained without slicing tokens, thus our model's vocabulary is exactly our artificial language's vocabulary.

\begin{table}[h]
 \begin{small}
 \begin{center}
 \begin{tabular}{ |l|c| } 
 \hline
  \textbf{Layers} & 3 \\
  \textbf{Attention Heads} & 4 \\
  \textbf{Hidden Size} & 256 \\
  \textbf{Intermediate Size} & 1024 \\
  \textbf{Training steps} & 300000 \\
 \hline
 \end{tabular}
 \caption{\label{tab:hyperparameters} Hyperparameters of our tested models.}
 \end{center}
 \end{small}
 \end{table}

\section{Training Details}
Here we display the parameters used to train our models.
\begin{table}[h]
 \begin{small}
 \begin{center}
 \begin{tabular}{ |l|c| } 
 \hline
    \textbf{Weight Decay} & 0.01\\
    \textbf{Learning Rate} & 5e-5 \\
    \textbf{Batch Size} & 8 \\
    \textbf{Optimizer} & Adam \\
 \hline
 \end{tabular}
 \caption{\label{tab:training-parameters} Hyperparameters used to train our models.}
 \end{center}
 \end{small}
 \end{table}

\section{Model Perplexities}
\label{app:model-perplexities}
We display the perplexities reached by our models on our validation sets in \cref{tab:perplexities}. Note that these perplexities are obtained in the traditional masked language modelling setting, where only one word is considered to be the ground truth. This explains the discrepancy when compared to model cross-entropies in \cref{fig:model-vs-data-entropies}. Contrarily to the rest of our analysis, these perplexity scores do not take the true probability distribution of the task into account, as only one label gets all the probability mass.
 
 \begin{table}[h!]
 \begin{scriptsize}
 \begin{center}
 \begin{tabular}{ |l|cccccc| } 
  \hline
  \multirow{2}{*}{\textbf{Model}}  & \multicolumn{6}{c|}{\textbf{\# Masked Words}} \\
   & \textbf{1} & \textbf{2} & \textbf{3} & \textbf{4} & \textbf{5} & \textbf{6} \\
  \hline
	\textbf{BERT} & 15.12 & 17.06 & 17.6 & 19.02 & 20.24 & 20.37 \\
	\textbf{NP} & 20.14 & 41.93 & 55.45 & 70.64 & 93.58 & 107.46 \\
 \hline
 \end{tabular}
 \caption{\label{tab:perplexities} Perplexities reached by our tested models for varying numbers of masked words.}
 \end{center}
 \end{scriptsize}
 \end{table}
 

 
\section{Asymmetries in KL-divergences}
In \cref{fig:kldiv-model-data-results}, when increasing the amount of masking, the increase in $D_{KL}(p_{u}, q_{BERT})$ is greater than that of $D_{KL}(p_{o}, q_{NP})$. While this could look surprising at first glance, it could simply be due to the asymmetric nature of cross-entropy, driven by the non-nullity of its left argument. \textbf{BERT} is strongly penalized by the true probability of completions it has never seen in ordered contexts: for those, it should have close to zero probability. \textbf{NP} in turn is less penalized, because it should have a non-zero probability for any completion assuming ordered context, as these sets of words are possible completions assuming unordered contexts. 

\end{document}